\documentclass[a4paper]{article}

\usepackage{INTERSPEECH2019}

\title{Exploring Methods for the Automatic Detection of Errors in \\ Manual Transcription}
\name{Xiaofei Wang$^{1}$, Jinyi Yang$^{1}$, Ruizhi Li$^{1}$, Samik Sadhu$^{1}$, Hynek Hermansky$^{1,2}$}
\address{
  $^1$Center for Language and Speech Processing, The Johns Hopkins University, USA\\
  $^2$Human Language Technology Center of Excellence, The Johns Hopkins University, USA}
\email{xiaofeiwang@jhu.edu, hynek@jhu.edu}

\begin{document}

\maketitle
\begin{abstract}
Quality of data plays an important role in most deep learning tasks. In the speech community, transcription of speech recording is indispensable. Since the transcription is usually generated artificially, automatically finding errors in manual transcriptions not only saves time and labors but benefits the performance of tasks that need the training process. Inspired by the success of hybrid automatic speech recognition using both language model and acoustic model, two approaches of automatic error detection in the transcriptions have been explored in this work. Previous study using a biased language model approach, relying on a strong transcription-dependent language model, has been reviewed. 
In this work, we propose a novel acoustic model based approach, focusing on the phonetic sequence of speech. Both methods have been evaluated on a completely real dataset, which was originally transcribed with errors and strictly corrected manually afterwards. 
\end{abstract}
\noindent\textbf{Index Terms}: Transcription error detection, phoneme posteriors, forced alignment, biased language model.

\section{Introduction}

Current automatic speech recognition (ASR) systems are trained on a large amount of data, with speech recordings and corresponding transcriptions. Human transcribers are usually employed to write down texts while listening to the pre-recorded audios with the help of particular software or platforms. Obviously, there exists a trade-off between transcription efficiency and its accuracy \cite{hazen2006automatic}. On one aspect, skilled transcribers are in great demands so that the cost of employment is relatively high. On the other hand, even a highly-skilled transcriber is not able to guarantee a 100\% accuracy of the transcription. Therefore, it is necessary to have some automatic post-processing techniques to detect the errors in manual transcriptions and correct the errors to make sure that the dataset is accurate enough for the corresponding task. Since the errors are rare compared to the whole dataset, reliable automatic detection techniques can quickly help locate the errors, optimizing the cost of both labors and time \cite{barras2001transcriber}.

The most straight-forward way to detect the errors is using a well-trained ASR system to find the mismatch between the decoding results and the transcription \cite{tian2010speech}\cite{li2016model}. Since the recognizer is usually not perfect, judgment would not be always reliable. Apparently, the recognizer should be taken advantage of in a thorough manner.

In conventional hybrid ASR, acoustic model (AM) represents the relationship between an audio signal and the phonemes
that make up speech, and language model (LM) provides the context to distinguish between words and phrases that sound similar \cite{deng2013recent}. 
They both contribute to finding the best hypotheses given an audio signal and help to detect the errors. 
For language model, the biased LM, which was built based on individual utterance with some top frequent words occurred in the training set, can strongly bias the decoding results towards transcription, unless there are errors in the transcription \cite{peddinti2016far}\cite{yang2019towards}. 
For acoustic model, a Viterbi alignment can be derived based on the given transcription \cite{gubian2009novelty}. The mean and variance of the best-path log-likelihood were used as confidence measures to determine if the transcription has errors or not \cite{tanamala2017semi}. In this way, only the fact of transcription itself was considered, while a classification output is still helpful which might reflect the ground truth, according to our previous study \cite{wang2018stream}\cite{wang2019stream}.

In this work, while given the well-trained hybrid ASR system, speech recordings, and the corresponding transcriptions, we investigate both the linguistic and phonetic contributions to detect the errors in the transcription. The mismatch between the alignment and posteriors from the classifier is novelly utilized. The remainder of this paper is organized as follows: In Section 2, we describe the methods to detect the erroneous transcriptions. Section 3 defines the experiments and results and section 4 follows with the conclusion.

\section{Methods}

\begin{figure*}[t]
  \centering
  \includegraphics[width=\linewidth]{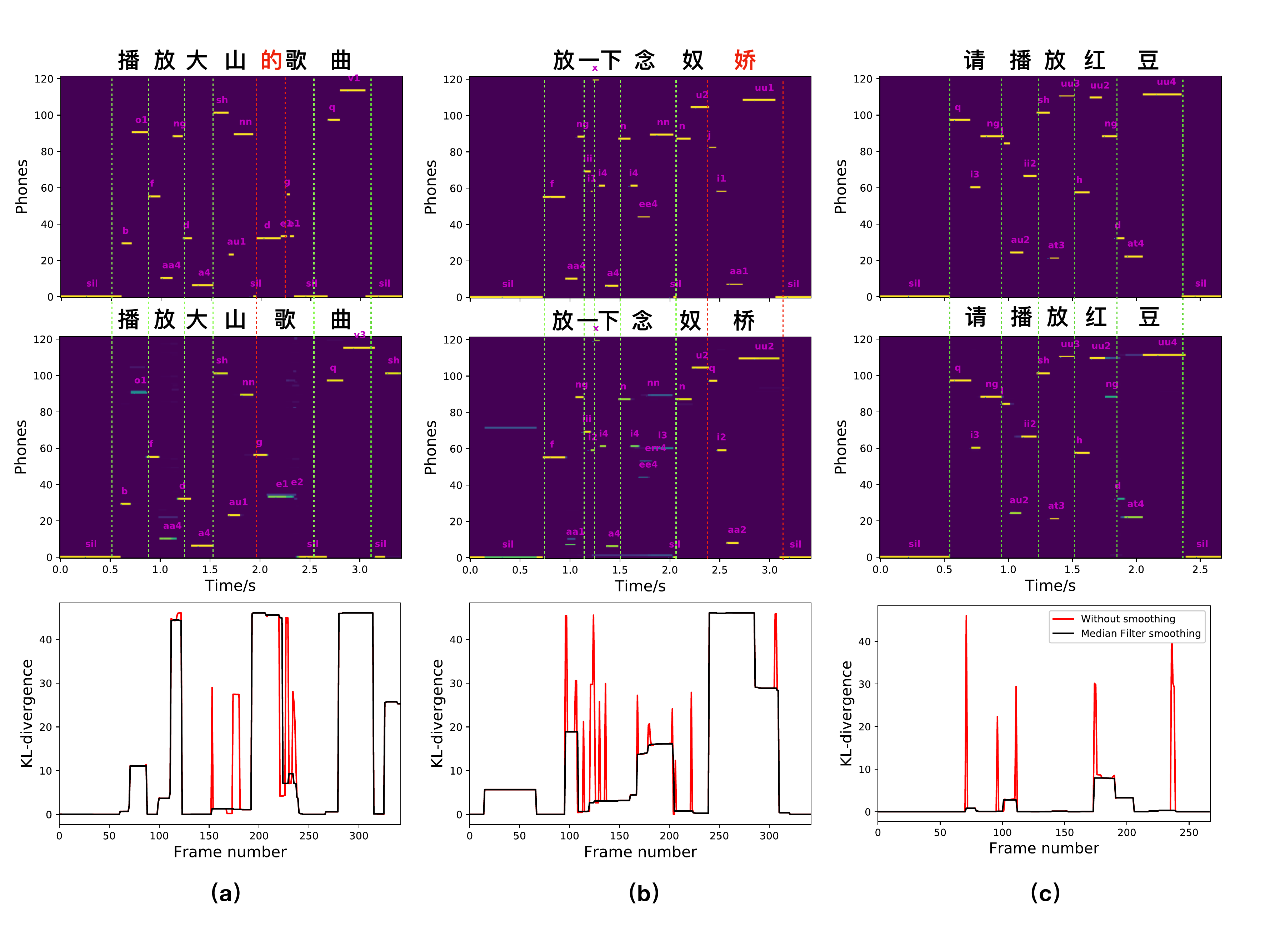}
  \caption{Posteriorgrams of forced alignment (top) and uniform phoneme graph (middle) as well as their computed KL-divergences (bottom). (a) Insertion transcription error; (b) Substitution transcription error; (c) Correct transcription. Note: the transcription errors are highlighted in red color.}
  \label{fig:1}
\end{figure*}

In this section, we describe two methods to detect the erroneous transcription, which are based on the language model and acoustic model, respectively. The former one emphasizes the word-level mismatch between the transcription and fully decoding results, and the latter one pays more attention to the phoneme-level information, which keeps the acoustic details of the speech.

\subsection{Using Language Model}
\subsubsection{Biased Language Model}
Biased language model trained on imperfect transcriptions has been used in lightly supervised training of the ASR system. The term ``biased'' refers to building a separate uni-gram LM for each utterance, with most words only from its own transcription and the most frequent words from training transcriptions. Intuitively, the prediction from the biased LM strongly agrees with ASR decoding result when manual transcription does not have errors. 

\subsubsection{Detection using biased LM}
Similar to our previous study, we decode the utterance using the corresponding biased LM, as well as a well-trained acoustic model, to generate a lattice instead of only choosing the best path \cite{peddinti2016far}. Given the lattice, the \textit{lattice oracle} word error rate is computed, which is the minimum Levenshtein edit distance between a path in the lattice and the transcription. It is expected to be low if the words in the transcription are seen in the lattice. 

\subsection{Using Acoustic Model}

\subsubsection{Phoneme based forced alignment}
Since we already have the transcription and a well-trained Hidden Markov Model (HMM) based hybrid ASR system, it is straightforward to get the alignment of each utterance, which is a representation of the sequence of HMM states. Alignment can be derived from aligning the reference transcript by using the Viterbi (best-path) approach. If focusing on the phoneme, we are able to merge context-dependent tri-phone state posteriors into monophone probabilities and perform forced-alignment at monophone level.

The phoneme-based forced alignment reflects the fact of the transcription, no matter whether the transcription is correct or not. It finally builds a relationship between the time $t$ and phonemes from transcription frame-wisely, which is given as follows:
\begin{equation}
\mathbf{p}_t = [ p^{(1)}_t, ... , p^{(k)}_t, ..., p^{(K)}_t]^T, k = 1,...,K
\end{equation}
, where $K$ is the number of phonemes and only one element $p^{(k)}_t$ in $\mathbf{p}_t$ equals $1$ due to the best-path property of the forced alignment, which is supervisely obtained. $[\star]^T$ is the transpose operation.

\subsubsection{Phoneme based posterior probabilities}
The posterior probabilities from the classifier always yield hypotheses based on the acoustic features frame-by-frame, regardless of what the transcription is. Even though errors might exist in the alignment, the hypotheses can reflect the ground truth if the acoustic model is good. 

For each frame, the posterior probabilities of the HMM tri-phone states can be obtained from the output layer of DNN based classifier, which uses soft-max non-linearity. Similar to the forced alignment, the posteriors of HMM states can be merged to phoneme posteriors according to the their mappings. Frame-based distribution of posteriors $\mathbf{q}_t$ is defined as follows:
\begin{equation}
\mathbf{q}_t = [ q(O_1|{\bf X}_t), ... , q(O_k|{\bf X}_t) , ..., q(O_K|{\bf X}_t) ]^T
\end{equation}
, where $q(O_k|{\bf X}_t), k = 1,...,K$ is the posterior probability of phoneme $O_k$ given the feature sequence ${\bf X}_t$ extracted from signals at time $t$.

In general, posteriors will sum to one but have very small values for most of the phonemes. Thus, to focus on the most significant phoneme sequences that the classifier predicts, a uni-gram phonotactic graph is built instead to generate the phone lattices, yielding a phoneme recognizer, with each phoneme having equal prior probability. Essentially, the phoneme recognizer plays a role in emphasizing the posteriors while pruning the less important sequences.

\subsubsection{Comparison between the forced alignment and classifier posteriors}

In the AM based approach, frame-based alignments are as the fact of transcription (may have errors) while posteriors are as hypotheses. 
Ideally, for each utterance, the transcription is correct when they reveal consistent sequences. 

To demonstrate the diversity of phonemes, we test the idea on Mandarin speech, which contains different tones for most of the phonemes. 
Fig.\ref{fig:1}(a) and Fig.\ref{fig:1}(b) are examples that the transcription has an \textit{insertion error} and a \textit{substitution error}, respectively. 
The upper two sub-figures represent probability distributions over the phone set sampled with 10 ms frame length, and show how likely a certain phone has been uttered at a given time.

In Fig.\ref{fig:1}(a), the forced alignment has an extra syllable {\bf"d e1"} (the red character, between 2.0s - 2.5s) due to the transcription error while the posteriors show the correct answer. In Fig.\ref{fig:1}(b), the last character of the speech recording (2.5s - 3.0s) is {\bf"q i2 aa2 uu2"}, but it was manually transcribed to be {\bf"j i1 aa1 uu1"}, which has been detected as well. Once we have a well-trained AM, it is feasible to detect such kind of errors in the transcriptions by comparing the difference between the alignment and posteriors.

\subsubsection{Kullback-Leibler (KL) divergence-based erroneous transcription detection}
The KL-divergence is a measure of how one probability distribution is different from the reference probability distribution. Given the forced alignment and phoneme posteriors, we calculate the frame-based symmetric KL divergence between them, shown as follows:
\begin{equation}
\mathcal{D}(\mathbf{p}_t, \mathbf{q}_t) = \sum_{k=1}^{K} p_t^{(k)}\log \frac{p_t^{(k)}}{q_t^{(k)}} + \sum_{k=1}^{K} q_t^{(k)}\log \frac{q_t^{(k)}}{p_t^{(k)}}
\end{equation}
, where $q_t^{(k)}=q(O_k|{\bf X}_t)$.

In general, the boundaries of each estimated phoneme are blurred and not well aligned. To solve this problem, we follow the strategy in \cite{burget2008combination}, which uses a soft alignment. At each frame $t$, a context of $2N+1$ frames ($[t-N, t+N]$) was used and the value of the center frame is weighted using a Median Filter, yielding a more smoothing output:
\begin{equation}
\mathcal{\hat{D}}(\mathbf{p}_t, \mathbf{q}_t) = {\bf Smooth}(\mathcal{D}(\mathbf{p}_{t'}, \mathbf{q}_{t'})), {t'} \in [t-N, t+N]
\end{equation}

The bottom sub-figures in Fig.\ref{fig:1} show the frame-by-frame non-smoothing and smoothing KL-divergence. The smoothing strategy overcomes the blurred boundary problem.
After smoothing, the peaks can be observed when the transcription has errors, while a correct transcription yields a small value range of KL-divergence, shown in Fig.\ref{fig:1}(c). In this work, we use the standard deviation (std) of the KL-divergence within the whole utterance as a confidence measure to determine if the utterance has transcription errors. A larger std results in a higher possibility to detect the errors.

\section{Experiment setup and Results}
\subsection{Experiment setup}

\begin{figure}[htb]
  \centering
  \includegraphics[width=0.7\linewidth]{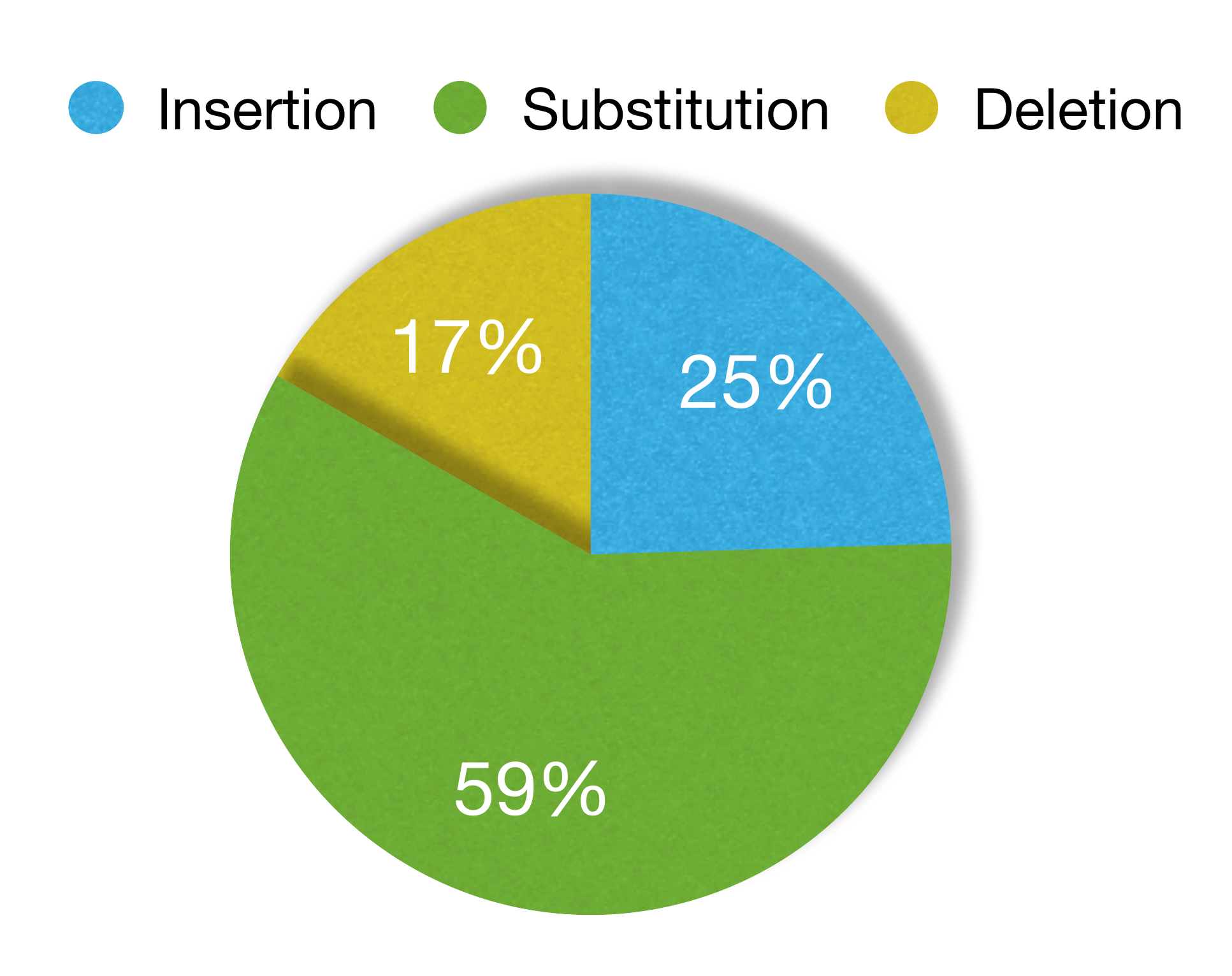}
  \caption{Distribution of the transcription errors. Compared to the correct transcriptions, total sentence error rate is $15.2\%$ and word error rate is $4.12\%$.}
  \label{fig:2}
\end{figure}

The evaluation set consists of 11,903 sentences (14.8 hours) which are Mandarin recordings with transcriptions manually edited by skilled transcribers. 1,810 sentences had transcription errors including Insertion ($1.01\%$), Deletion ($0.69\%$), and Substitution ($2.42\%$), shown by Fig.\ref{fig:2}. After an artificially correcting process  (re-listen the audios and check the transcriptions artificially), all the errors had been strictly modified to match the audios. Therefore, we got ground truths, which can be used to evaluate automatic detection schemes. Speech recordings, sampled at 16kHz, are close-talking smart-device commands and daily conversation sentences, acquired using high fidelity microphones. In this paper, 708.2 hours of training data covering the topics of evaluation set had been used to train the hybrid ASR system for the purpose of detecting the erroneous transcriptions.

The experiments were performed using the Kaldi speech recognition toolkit \cite{povey2011kaldi}. A triphone HMM-GMM system with speaker-adaptive training was used to generate the alignments and train the neural network acoustic model. The HMM was trained on 122 position dependent phonemes, with 3 silence phonemes and 119 non-silence phonemes. Time delay neural network (TDNN) based AM and general tri-gram word LM had been trained using 40-dimension MFCC \& 100-dimension i-vectors and lexicon from training data, respectively. We tested both evaluation sets with and without transcription errors using the well-trained ASR system. Table \ref{tab:wer} shows the word error rate (WER) of the evaluation sets with and without transcription errors. The transcription errors lead to a slight performance degradation because they supplied wrong references for ASR.

\begin{table}[!h]
  \caption{Word Error Rate (WER\%) on hybrid ASR system consisting of AM and general tri-gram word LM.}
  \label{tab:wer}
  \centering
  \begin{tabular}{ c|c|c }
    \toprule
    \toprule
    {\textbf{Model/WER(\%)}} & {\textbf{w trans. errors}} & {\textbf{w/o trans. errors}} \\
    \midrule
    HMM-GMM                       & 15.94 & 15.18           \\
    HMM-TDNN                      & 11.59 & 10.67              \\
    \bottomrule
    \bottomrule
  \end{tabular}
\end{table}

\subsection{Forced alignment and phoneme posteriors}

Phoneme-based forced alignment of each utterance was generated by the HMM-GMM AM. It was converted to posteriors where each frame has a single phoneme with unit posterior.
We used a uniform phonetic LM and the HMM-TDNN AM to get the phoneme posteriors.
As for the KL-divergence calculation, we used a 15-frame ($N=7$) Median filter to calculate the smoothed score between the alignment and phoneme posteriors.

\subsection{Biased ASR system}
We followed the procedures that were used to build the biased ASR system utterance-by-utterance \cite{peddinti2016far}\cite{yang2019towards}. Firstly, a four-gram unmodified Kneser-Ney interpolated language model from the transcript for each utterance was built. 
This language model was then interpolated with a uni-gram language model estimated using counts of the 100 most frequent words in the whole training data set. 
Note that the uni-gram language model allows the decoding process to predict word sequences that are not the same as the transcription, therefore the decoded lattice is more likely to include paths that are close to speech.

\subsection{Results}

\begin{figure}[t]
  \centering
  \includegraphics[width=\linewidth]{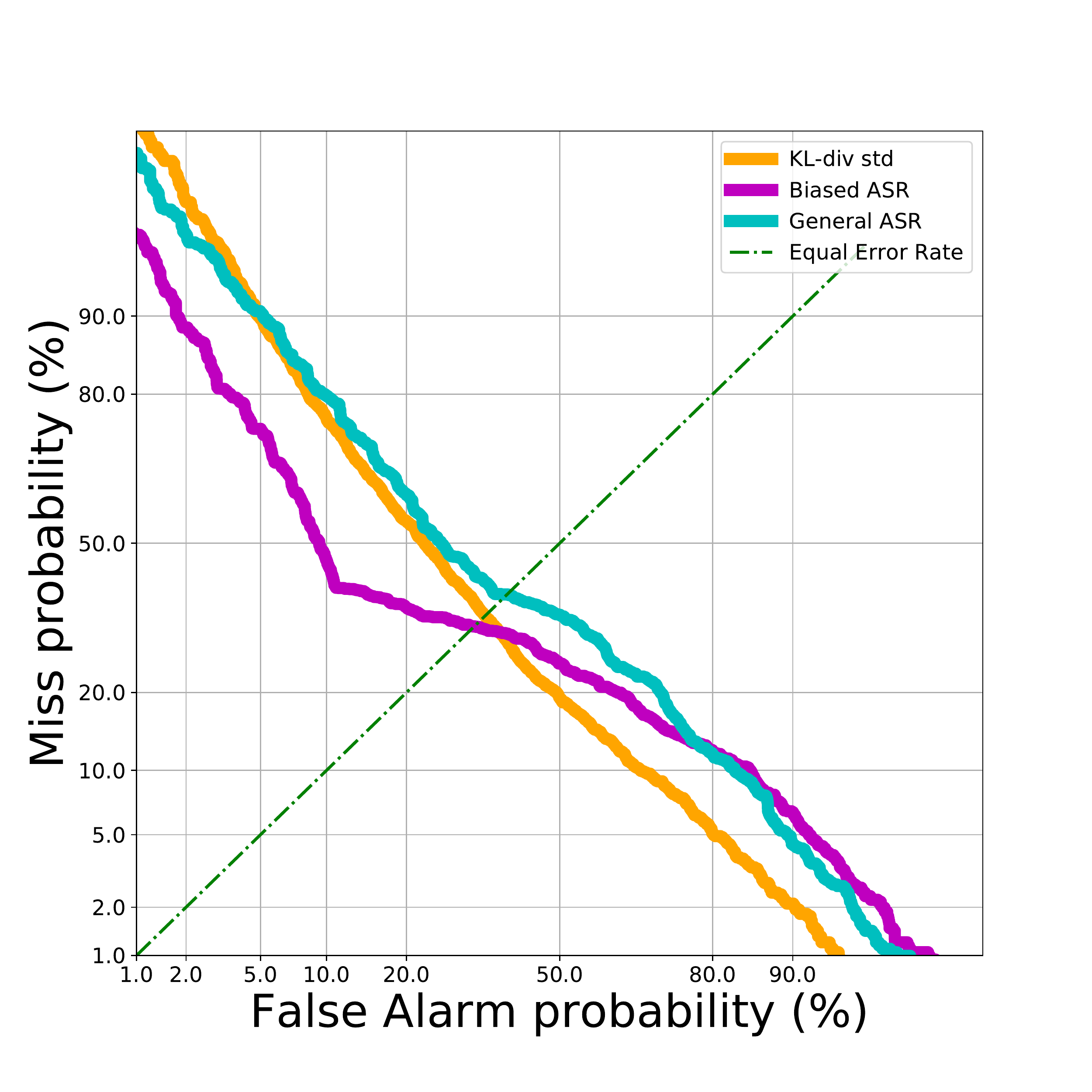}
  \caption{The DET curves of different approaches in detecting the transcription errors.}
  \label{fig:3}
\end{figure}

\begin{table}[!h]
  \caption{Equal Error Rate (EER\%) using different approaches.}
  \label{tab:eer}
  \centering
  \begin{tabular}{ c|c|c|c }
    \toprule
    \toprule
    {\textbf{Method}} & {\textbf{General ASR}} & {\textbf{Biased LM}} & {\textbf{KL-div std}} \\
    \midrule
    {\textbf{EER(\%)}}      & 38.62 & 31.95 & 34.11         \\
    \bottomrule
    \bottomrule
  \end{tabular}
\end{table}

The Detection Error Tradeoff (DET) curve has been used to derive the Equal Error Rate (EER), which is the point where false positive rate is equal to false negative rate. Lower ERR results in better performance. DET curves of three compared approaches are derived in Fig.\ref{fig:3} using the following setups.

{\bf General ASR}: As a baseline, we decode each utterance in the evaluation set using the HMM-TDNN AM and tri-gram word LM. WER of each utterance is compared with a threshold ranged from 0 to 1.

{\bf Biased ASR}: Each utterance is decoded using HMM-TDNN AM and biased LM. WER of each utterance is compared with a threshold ranged from 0 to 1. 

{\bf KL-div std}: Standard deviation of the smoothed KL-divergence for each utterance is calculated. The std score is compared with a threshold ranged from 0 to 20.

The Biased ASR and KL-div std show comparable performance from the DET curves in Fig .\ref{fig:3}. The EERs of them are $31.95\%$ and $34.11\%$, respectively, which are better than the baseline general ASR approach (Table \ref{tab:eer}). One advantage of the proposed KL-div std method is that it is much more efficient, without doing the full-pass decoding as the other ones did, which is helpful for a huge amount of data.

According to the convex direction, Fig.\ref{fig:3} also illustrates how to select the detection scheme given a particular task.
If we would like to find more erroneous transcriptions to make sure the total transcription accuracy of the dataset can achieve some numbers (usually over 95\%-98\% for commercial uses), KL-div std would be the best choice. While biased LM approach is desirable if we want to have higher recall accuracy.

The problem of transcriptions from Mandarin speech is that there exist many homonyms. To the best of our knowledge, both methods cannot deal with this kind of problem, which made the EER higher than other languages.

\section{Conclusions}
The work presented two alternative strategies to automatically detect the errors in the manual speech transcription. From a linguistic view, we used the biased language model to find the difference between the hypotheses and transcription. In addition, we proposed to use mismatch between the forced alignment and posteriors of the acoustic classifier to detect the phonetic errors, which is more intuitive. Facing a completely real Mandarin dataset, in which manual transcription errors reasonably distributed, these two utterance-based measures showed comparable ability in detecting the transcription errors.
Since the perspectives of these two measures are complementary, which are respectively from the linguistic and phonetic view of speeches, finding a combination of two would be interesting in the future. Possible extensions of the work towards correcting pronunciation errors would be interesting as well.
%Exploiting the information from transcription, we derived the phoneme-based forced alignment for each utterance. From the views of linguistic and phonetic of speech signal,  
%By comparing the hypotheses decoded by using the biased language model and using the acoustic classifier
%the transcription itself and acoustic forced alignments were utilized as the references, which may tell different stories from the hypotheses decoded by using the biased language model and using the acoustic classifier, respectively. 
%Evaluated on a completely real dataset where all the transcription errors were generated accidentally during transcribing, these two utterance-based measures showed equal importance in detecting the transcription errors.

\section{Acknowledgements}
This work is supported by a gift to Hynek Hermansky from Beijing Magic Data Technology Co., Ltd., a Google faculty award to Hynek Hermansky and National Science Foundation under Grant No. 1704170. We would like to thank Beijing Magic Data Technology Co., Ltd. for providing the real dataset for this work.

\bibliographystyle{IEEEtran}
\bibliography{mybib}

\end{document}